\documentclass[acmtog,nonacm]{acmart}

\usepackage[mathletters]{ucs}
\usepackage{soul}
\usepackage[utf8]{inputenc}

\usepackage[ruled]{algorithm2e} %

\usepackage{boldline} %
\usepackage{multirow}
\usepackage{stfloats}
\usepackage{float}

\providecommand{\F}{}

\providecommand{\N}{}

\providecommand{\R}{}
\providecommand{\S}{}

\providecommand{\b}{}

\providecommand{\l}{}

\providecommand{\r}{}

\providecommand{\v}{}

\providecommand{\x}{}

\usepackage{tabularx}
\usepackage{booktabs}
\usepackage{makecell}
\definecolor{white}{rgb}{1,1,1}
\definecolor{lightbluishgrey}{rgb}{0.76471,0.84824,0.91647}

\usepackage{subcaption}
\usepackage{wrapfig}
\usepackage{graphicx}

\usepackage{listings}
\usepackage{layouts}
\makeatletter 
\newcommand{\layoutdetails}{%
\begin{tabular}{ll}
 \texttt{\textbackslash{textwidth}} & \printinunitsof{in}\prntlen{\textwidth} \\
\texttt{\textbackslash{linewidth}} & \printinunitsof{in}\prntlen{\linewidth} \\
Main text font &  \f@size pt \f@family \\
\sffamily \small Caption text font &  \sffamily \small \f@size pt \f@family \\
\end{tabular}%
}
\makeatother

\newcommand{\rev}[1]{{#1}}

\renewcommand{\x}{\mathbf{x}}
\renewcommand{\b}{\mathbf{b}}
\renewcommand{\l}{\mathbf{l}}
\renewcommand{\r}{\mathbf{r}}
\renewcommand{\v}{\mathbf{v}}
\renewcommand{\F}{\mathbf{F}}

\renewcommand{\N}{\mathcal{N}}
\renewcommand{\R}{\mathbb{R}}
\renewcommand{\S}{\mathbb{S}}

\begin{document}

\title{2D Neural Fields with Learned Discontinuities}

\author{Chenxi Liu}
\orcid{0000-0003-3613-1662}
\affiliation{%
  \institution{University of Toronto}
 \city{Toronto}
 \country{Canada}}
\email{chenxil@cs.toronto.edu}

\author{Siqi Wang}
\orcid{0000-0001-9722-0314}
\affiliation{%
 \institution{New York University, Courant Institute of Mathematical Sciences}
 \city{New York}
 \country{USA}}
\email{siqi.wang@nyu.edu}

\author{Matthew Fisher}
\orcid{0000-0002-8908-3417}
\affiliation{%
 \institution{Adobe Research}
 \city{San Francisco}
 \country{USA}}
\email{matfishe@adobe.com}

\author{Deepali Aneja}
\orcid{0000-0001-9610-5648}
\affiliation{%
 \institution{Adobe Research}
 \city{Seattle}
 \country{USA}}
\email{aneja@adobe.com}

\author{Alec Jacobson}
\orcid{0000-0003-4603-7143}
\affiliation{%
 \institution{University of Toronto}
 \city{Toronto}
 \country{Canada}}
\email{jacobson@cs.toronto.edu}
\affiliation{%
 \institution{Adobe Research}
 \city{Toronto}
 \country{Canada}}
\email{alecjacobson@adobe.com}

\begin{abstract}
  
Effective representation of 2D images is fundamental in digital image processing, where traditional methods like raster and vector graphics struggle with sharpness and textural complexity respectively.
Current neural fields offer high-fidelity and resolution independence but require predefined meshes with known discontinuities, restricting their utility.
We observe that by treating all mesh edges as potential discontinuities, we can represent the magnitude of discontinuities with continuous variables and optimize.
Based on this observation, we introduce a novel discontinuous neural field model that jointly approximate the target image and recovers discontinuities.
Through systematic evaluations, our neural field demonstrates superior performance in denoising and super-resolution tasks compared to InstantNGP, achieving improvements of over 5dB and 10dB, respectively.
Our model also outperforms Mumford-Shah-based methods in accurately capturing discontinuities, with Chamfer distances $3.5\times$ closer to the ground truth.
Additionally, our approach shows remarkable capability in approximating complex artistic and natural images and cleaning up diffusion-generated depth maps.

\end{abstract}



\begin{teaserfigure}
  \includegraphics[width=\linewidth]{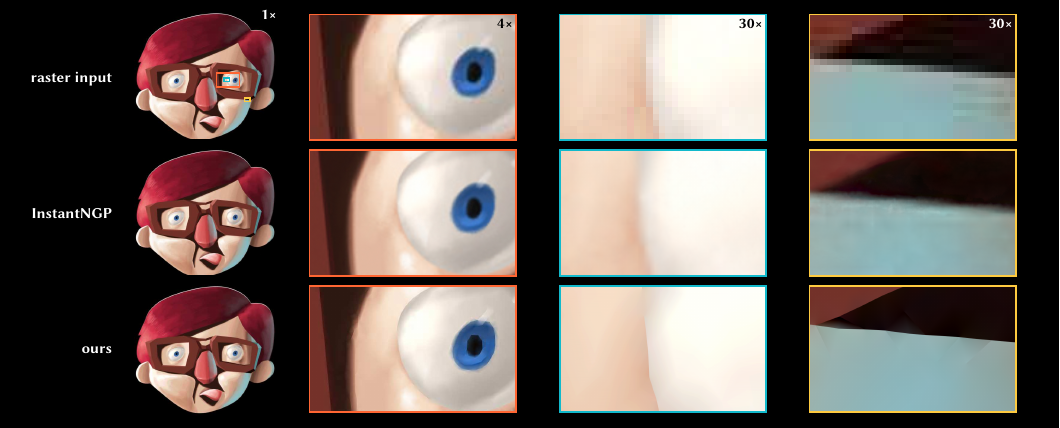}
  \caption{
    We propose a discontinuous 2D neural field that can jointly approximate the target image and recover unknown discontinuities.
    Our neural field supports edge-preserving denoising and arbitrary resolution while continuous neural field, such as InstantNGP \cite{muller22}, blurs region boundaries under close view.
    Zoom scales indicate inference scale. Zoom in to see details.
  }
  \label{fig:teaser}
\end{teaserfigure}

\maketitle

\section{Introduction}

Digital image representations --- such as pixel arrays or vector graphics --- discretize \emph{image functions} that map 2D locations to colors.
For a variety of reasons (occlusions in captured 3D scenes, layers in graphic designs, material boundaries, etc.), typical image functions are well modeled as continuous functions \emph{almost everywhere} with sparse discontinuities appearing along 1D curves.
Unfortunately, images stored as regular grids of pixel colors do not directly model discontinuities and assume a fixed resolution.
Meanwhile, vector graphics formats (e.g., \texttt{.svg}) represent resolution-independent discontinuities directly with curves, fill boundaries, or layer overlaps, but these formats have minimal support for continuous signals elsewhere (e.g., solid colors, basic gradients).

\begin{figure}[t]
  \includegraphics[width=\linewidth]{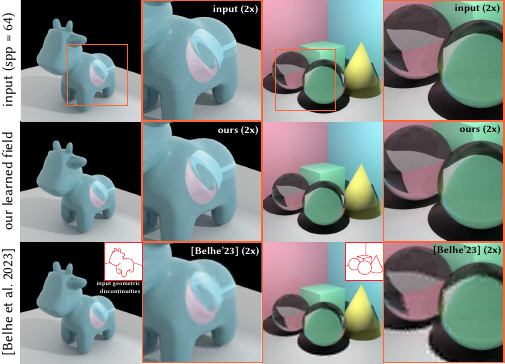}
  \caption{
    Discontinuity-aware 2D neural field \cite{belhe23} requires accurate 2D discontinuities as input.
    In their example of denoising 3D renderings, all types of discontinuities may not always be available. False negatives caused by sharp texture and refracted geometries lead to blurs.
  }
  \label{fig:belhe_comp}
\end{figure}

Recently, \citet{belhe23} proposed storing images as the output of a small neural network fed a spatially varying feature vector.
That feature vector is carefully interpolated over a triangle mesh constructed to ensure discontinuities along certain edges.
The weights of the neural network are optimized to reconstruct samples of an image function.
This exciting representation is very compact and especially well suited for noisy input samples.
Unfortunately, discontinuities must be given in advance as input, and the function space used for interpolation by \citeauthor{belhe23} does not immediately make it clear how to treat discontinuity locations as optimization variables.
When discontinuties are missing --- even partially -- their reconstruction is noticeably poor (see Fig.~\ref{fig:belhe_comp}).

In this paper, we propose a non-trivial change to \citeauthor{belhe23}'s method so that image fitting no longer requires discontinuities to be known in advance.
During fitting, we treat \emph{all} mesh edges as potential discontinuities, introducing variables to model the magnitude of value-jump across the edge.
These variables are \emph{continuous} and happily optimized along with feature vectors and mesh vertex positions during gradient-based reconstruction.
Unlike \citeauthor{belhe23}'s function space, our function space easily affords a post-processing procedure to identify and merge almost-continuous edges,
\rev{greatly improving storage efficiency while maintaining reconstruction fidelity.}

We demonstrate that our neural fields with learned discontinuities directly support denoising and super-resolution.
Using a novel systematically synthesized diffusion curve dataset, we show that our method outperforms InstantNGP \cite{muller22} with matching size by $>5$dB in the denoising task and $>10$dB when examined under super-resolution.
Visually, our neural field maintains sharp region boundaries at large zoom levels ($30\times$ in Fig.~\ref{fig:teaser},\ref{fig:sec2_comp},\ref{fig:dc_denoise}) while InstantNGP blurs boundaries.
Our method recover more accurate discontinuities than Mumford-Shah-based denoising \cite{wang22}: $3.5\times$ smaller Chamfer distance to the groundtruth.
We show that our neural fields can approximate typical vector graphics images corrupted by JPEG compression.
The use cases of our method also include general 2D data, such as diffusion-generated depth maps, which our method segments with clear cuts between depth discontinuities.
Finally, we stress test our method with complicated artistic drawings and natural images (Fig.~\ref{fig:teaser},\ref{fig:vec}).

\section{Related Work}

\begin{figure*}
  \includegraphics[width=\linewidth]{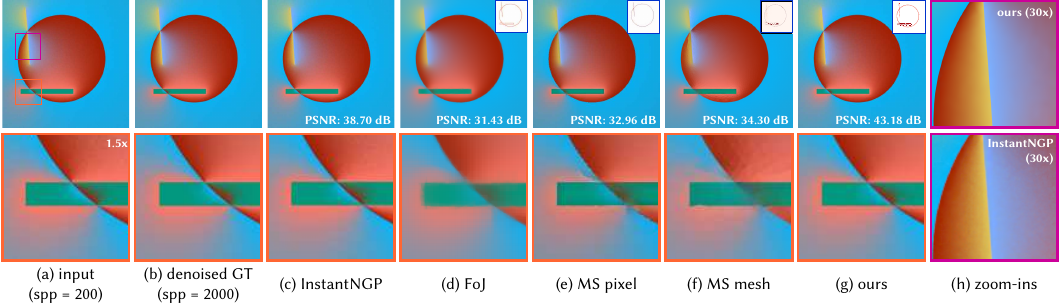}
  \caption{
    (a,b) Diffusion curves define an example harmonic function field with sharp discontinuities \cite{orzan08}.
    (c) Continuous neural fields, such as InstantNGP \cite{muller22}, do not represent discontinuities, resulting in blur image when zoomed-in.
    (d) Denoising method, Field of Junctions (FoJ) \cite{verbin21}, fails to recover clear discontinuities due to using constant-patch-based approximation.
    (e, f) Mumford-Shah functional based methods jointly approximate the target and detect discontinuities (see red edges in insets) \cite{wang22}, similar to our method.
    However, both versions fail to achieve both goals because of limited function expressiveness.
    (g) Our accurate approximation and recovered discontinuities.
  }
  \label{fig:sec2_comp}
\end{figure*}

\paragraph{Geometric representation for images}

Geometric image representations, such as vector graphics, address raster image limitations by encoding discontinuities as shapes with simple color functions. Methods combining accurate geometric boundaries with interior samples focus on representation design, interpolation, and real-time rendering \cite{bala03, tumblin04, sen04, ramanarayanan04, tarini05, parilov08, pavic10, reshetov16}.
Another approach represents digital images with discrete curves or region boundaries and smooth interior functions \cite{tian22}.
Diffusion curves \cite{orzan08} defines images as harmonic functions with curve Dirichlet boundary conditions, a space our method can accurately approximate (Section~\ref{sec:eval}).
Triangle mesh-based representations \cite{davoine96, demaret06, su04, tu11} and more advanced \rev{curve- and} patch-based primitives \cite{lecot06, sun07, xia09, lai09, zhao17} are introduced for vectorization of natural images.
These approaches, while similar to ours in merging discrete geometries and functional representations for interiors, often construct geometric boundaries separately, relying on edge detection, segmentation, or user input.
In contrast, our method jointly optimizes discontinuity locations and interior colors using a mesh without predefined cuts.

\paragraph{Neural fields}
Neural fields are widely used for representing spatial functions like images and signed distance fields (SDFs), often parameterized by neural networks \cite{xie22}.
Early work by \citet{song15} introduced coordinate neural networks for image encoding, a technique applicable to large image compression \cite{mildenhall21, sitzmann20, mehta21}.
Although these methods capture high-frequency details, they often fail to represent true discontinuities, causing blurring at high zoom levels.

Hybrid neural fields, or feature fields, combine neural networks with discrete data structures like grids \cite{chen21, martel21, takikawa21, yu21, shen21, muller22, chen22, takikawa23}.
These offer reduced computation, better network capacity utilization, and explicit geometric representation. While most hybrid fields don't explicitly address discontinuities, recent methods like ReLU fields \cite{karnewar22} attempt to approximate them with steep ReLUs, but still face blurring issues as shown by \citet{belhe23}.
Other approaches, like those by \citet{reddy21}, represent true discontinuities but are tailored for specific cases like fonts. Discontinuity-aware 2D neural fields \cite{belhe23}, an inspiration for our approach, show promise but require user-provided discontinuous edges (Fig.~\ref{fig:belhe_comp}).
Our method fits unknown discontinuities and is compatible with discontinuity-aware 2D neural field offering efficient storage and inference.

Concurrent work also uses supervised learning to generate a field of patches for discontinuity detection, especially in noisy environments \cite{polansky24}.
We compare to its preceding work \cite{verbin21} and show that their focus is on denoising rather than accurate approximation (Fig.~\ref{fig:sec2_comp}d).

\paragraph{Differentiable rendering}
Addressing visual discontinuity in differentiable rendering is a persistent challenge \cite{zhao20, spielberg23}. 
Traditional automatic differentiations often overlook discontinuities' contributions to gradients \cite{bangaru21}. Differentiable rendering techniques include smoothed rasterization \cite{loper14, kato18, liu19, laine20},
Monte Carlo-based methods with boundary and interior sampling \cite{li18, li20, loubet19, bangaru20, zhang23}, 
analytic solutions \cite{bangaru21}, 
and finite difference approaches \cite{yang22, deliot24}. 
Our approach introduces a differentiable neural primitive capable of representing discontinuities. 
By framing our approximation problem as an inverse rendering task, our method efficiently handles discontinuities by leveraging existing differentiable rendering techniques.

\paragraph{Mumford–Shah functional}
The Mumford–Shah (MS) functional, proposed for image segmentation \cite{mumford89}, models images as piecewise-smooth functions with explicit discontinuities.
The Ambrosio-Tortorelli approximation \cite{ambrosio90} made solving for MS functionals tractable with techniques like ADMM.
This functional has been widely used in 2D image tasks \cite{tsai01, vese02, le22}
and applied to 3D surfaces \cite{tong16, bonneel18, wang22} and volumes \cite{coeurjolly16}.
Our method, similar to the MS functional, jointly recovers piecewise-smooth functions and discrete discontinuities.
Unlike level-set-based methods \cite{vese02, esedoglu02}, ours supports open boundaries and produces more accurate discontinuities free of narrow bands \cite{tong16, bonneel18} and staircasing artifacts \cite{tsai01, le22}.
While \citet{wang22} also discretize discontinuities on mesh edges, their representation (constant colors per face) lacks the expressiveness of our neural model (Fig.~\ref{fig:sec2_comp}).

\section{Method}
\label{sec:method}

\begin{figure*}[t]
  \includegraphics[width=\linewidth]{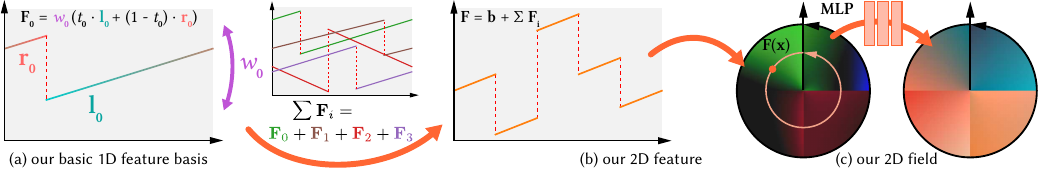}
  \caption{
    (a) Our 1D feature basis functions fit discontinuous target function with continuous variables \rev{($w, \l, \r$)}.
    (b) In a 2D vertex neighborhood, our 1D feature basis functions are defined along the radial direction.
    (c) The features have a constant piecewise slope per dimension, which is enhanced by an MLP.
  }
  \label{fig:feature_overview}
\end{figure*}

Given a 2D image function $I(\x), \x \in \R^2$, we aim to approximate it with a 2D neural field that is continuous everywhere, except for locations with sharp changes in the input.
In our implementation, $I(\x)$ is given by nearest-neighbor sampling of a raster image.

The desired properties of such 2D neural fields are formulated by \citet{belhe23}.
Let $\Omega$ be a 2D domain and $\Gamma = \{ \gamma_0, \cdots, \gamma_{n-1} \mid \gamma_i \in \Omega, \forall i\}$ be curves that can only intersect with each other at endpoints $\partial \Gamma$.
The directional discontinuity is defined with respect to a point $\x \in \R^2$, a polar coordinate system centered at $\x$ and a angular coordinate $\alpha \in \R$.
The \emph{directional limit} of a function at a position $\x$ along a direction $\theta$ is defined as $h(\x, \theta) = \lim_{r\rightarrow 0^+} f(\mathcal{C}(r, \theta))$, where $\mathcal{C}(r, \theta)$ maps the polar coordinate to Cartesian coordinate (see $\x_0$ in inset).
A function is directionally discontinuous at $\x$ and $\alpha$ if
\begin{align}
  \lim_{\theta \rightarrow \alpha^+} h(\x, \theta) = \lim_{\theta \rightarrow \alpha^-} h(\x, \theta).
\end{align}

\begin{wrapfigure}{r}{0.3\linewidth}
\includegraphics[width=\linewidth,trim=0.9cm 0.7cm 0 0.9cm]{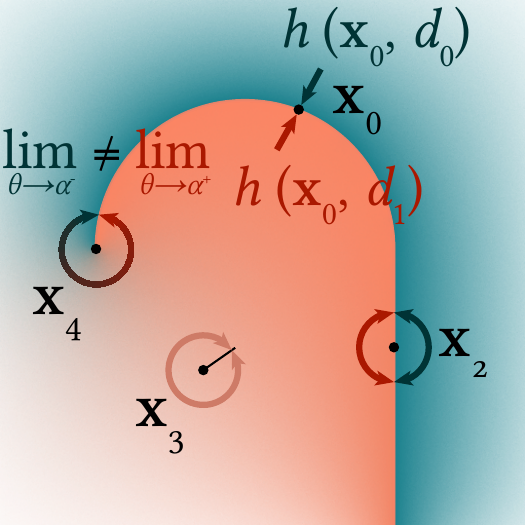}
\end{wrapfigure}

Following this continuity criteria, a field $f: \Omega \rightarrow \R^d$ is:
(1) continuous at $\x \in \Omega \setminus \Gamma$ (e.g., $\x_3$ in inset);
(2) directionally discontinuous for the two tangent directions at $\x \in \Gamma \setminus \partial \Gamma$ (e.g., $\x_2$);
(3) directionally discontinuous at the tangent direction pointing inwards to the curve(s) at $\x \in \partial \Gamma$ (e.g., $\x_4$).
Note that $\gamma_i$ can be a curve \cite{belhe23}. 
For simplicity, we only consider the case where $\gamma_i$ is a line segment and approximate curve geometries by adjusting triangle density.

\paragraph{Overview}
Our neural field is built upon an underlying triangle mesh and discontinuous feature function defined locally within each vertex one-ring.
To see how it works on triangle meshes, we first analyze the 1D case then extend the formulation to 2D (Section~\ref{sec:feature}).
This discontinuous feature function allows discontinuity across the mesh edges from which we learn the target discontinuous edge set $\Gamma$.
We approximate the target image by initializing a triangle mesh (Section~\ref{sec:init}), fitting a neural field with all edges being potentially discontinuous (Section~\ref{sec:opt}), and refining to enforce continuity on almost-continuous edges (Section~\ref{sec:round}).

\subsection{Neural Field with Discontinuous Features}
\label{sec:feature}

Given a triangle mesh in $\R^2$, we define learnable discontinuous feature function basis within vertex one-rings.
Optimization of discontinuous functions would typically require discrete operations, challenging automatic differentiation.
We observe that by treating all mesh edges as being \emph{potentially discontinuous}, we can represent the magnitude of discontinuity with \emph{continuous} variables.
In this way, we can optimize the continuous variables with the standard autodiff-gradient-based approach.
We first introduce our feature function basis in $\R$ (Section~\ref{sec:1d}) then extend it to $\R^2$ (Section~\ref{sec:2d}).

\subsubsection{Discontinuous field in 1D}
\label{sec:1d}

\begin{wrapfigure}{r}{0.2\linewidth}
\includegraphics[width=\linewidth,trim=0.7cm 0.5cm 0 0.8cm]{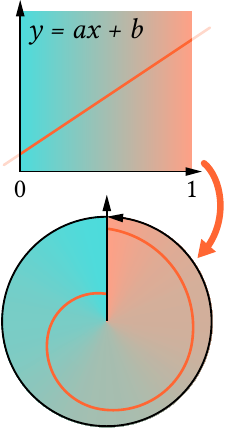}
\end{wrapfigure}
Consider a simple 1D linear function $g(x) = ax + b$.
We can restrict its domain to $x \in [0, 1)$ and map this function into the unit circle $\S^1$.
In this way, we construct a discontinuity at $x=0$ when $a \neq 0$.
This construction enables standard autodiff to calculate the left and right derivatives of $g(x)$ at the discontinuity location.
Given a closed polyline $S=(V, E)$ and an arbitrary reference coordinate $x \in \S^1$ on it, we define a local coordinate $x_i = t_i(x)$ centered at each vertex $v_i \in V$ and a linear function $\varphi_i(x_i) = a_i x_i + b_i$.
These functions form a basis for piecewise linear functions
\begin{equation}
  g(x) = \sum_{v_i \in V} \varphi_i(t_i(x)) = \sum_{v_i \in V} \left( a_i t_i(x) + b_i \right),
\end{equation}
which is discontinuous at $v_i$ if $a_i \neq 0$.
Note that its discontinuous pieces all share the same slope of $\sum_{v_i \in V} a_i$ (see Fig.~\ref{fig:feature_overview}a).

This single-value basis has limited expressiveness.
To improve this, we extend to $k$-dimensional features and over-parameterize.
\begin{align}
\F(x) &= \b + \sum_{v_i \in V} \F_i(t_i(x)) = \b + \sum_{v_i \in V} w_i \left(t_i(x) \cdot \l_i  + (1 - t_i(x)) \cdot \r_i \right) \label{eq:interp1d} \\
&= \b + \sum_{v_i \in V} \left( w_i (\l_i - \r_i) t_i(x) + w_i \r_i \right)
\end{align}
where, intuitively, $w_i \in \R$ controls whether a discontinuity is eliminated; $\l_i, \r_i \in \R^k$ are left and right features defined on either side of a vertex $v_i$; the last term in Eq.~\ref{eq:interp1d} defines a linear interpolation between these two features; $\b \in \R^k$ is a global bias (Fig.~\ref{fig:feature_overview}a).

We pass feature $\F(x)$ to an MLP \rev{($\R^k \mapsto \R^3$)} for our neural field
\begin{equation}
  f(x) = \text{MLP}(\F(x)).
\end{equation}
Our neural fields use shallow MLPs: two layers of 64 neurons for 1D, two layers of 128 neurons for 2D, with tanh activations.

\begin{wrapfigure}{r}{0.29\linewidth}
\includegraphics[width=\linewidth,trim=0.7cm 0.5cm 0 0.6cm]{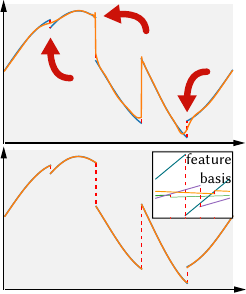}
\end{wrapfigure}

We conduct a simple 1D fitting experiment (inset).
The target function (blue) is a function of translated sine pieces and discontinuities at four points.
For reference, we construct a 1D feature field ($k=5$) with features defined at these four locations and linearly interpolated between them (orange,top).
We compare this reference to our 1D model ($k=2$) with potential discontinuities at the matching points (orange,below\rev{; feature basis functions are $\F_i$ in Eq.~\ref{eq:interp1d}}).
Contracted with our faithful approximation, the reference field not only fits the discontinuities (red dashed lines) with inaccurate steep continuous jumps but also fails to stay close to the continuous pieces.

\subsubsection{Discontinuous field in 2D}
\label{sec:2d}

\begin{wrapfigure}{r}{0.25\linewidth}
\includegraphics[width=\linewidth,trim=0.7cm 0 0 0.55cm]{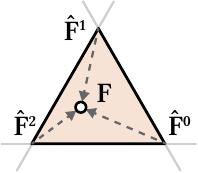}
\end{wrapfigure}
We extend this 1D field into 2D domain $\Omega$ on a triangle mesh $M=(V, T)$, where $V, T$ are vertex and face set.
The feature value at a point $\x \in \Omega$ intersecting a triangle $T_i = (v_i^0, v_i^1, v_i^2)$ is determined by the barycentric interpolation of three per-vertex local features
\begin{align}
  \F(\x) = (1 - \lambda_1 - \lambda_2)\hat{\F}_i^0(\x) + \lambda_1 \hat{\F}_i^1(\x) + \lambda_2 \hat{\F}_i^2(\x), \lambda_1, \lambda_2 \in [0, 1] \label{eq:bary},
\end{align}

The local features are constructed similarly to the 1D feature introduced in Section~\ref{sec:1d}.
Given a vertex $\v_i$, the local feature is
\begin{align}
  \hat{\F}_i(\x) &= \b_i + \sum_{\v_j \in \N(\v_i), i\neq j} \F_{i,j}(t_i^j(\x)) \\
  &= \b_i + \sum_{\v_j \in \N(\v_i), i \neq j} w_{i,j} \left(t_i^j(\x) \cdot \l_{i,j}  + (1 - t_i^j(\x)) \cdot \r_{i,j} \right) \label{eq:2d_interp} \\
  &= \b_i + \sum_{\v_j \in \N(\v_i), i \neq j} \left( w_{i,j} (\l_{i,j} - \r_{i,j}) t_i^j(\x) + w_{i,j} \r_{i,j} \right).
\end{align}

\begin{wrapfigure}{r}{0.27\linewidth}
\includegraphics[width=\linewidth,trim=0.72cm 0.55cm 0 0.55cm]{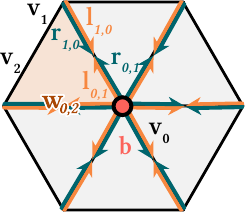}
\end{wrapfigure}
Here $\N(\v_i)$ is $\v_i$ one-ring and $\v_i, \v_j$ are connected by an edge;
$\b_i \in \R^k$ is a bias defined at the center $\v_i$;
$t_i^j(\x)$ maps $\x$ to the local polar coordinate system centered at $\v_i$ with $(\v_i, \v_j)$ as the polar axis.
To make the basis in Eq.~\ref{eq:2d_interp} a linear interpolation, we normalize $t_i^j(\x)$ from $[0, 2\pi)$ to $[0, 1)$.
Note that the bias $\b_i$ is introduced so a vertex reduces to a regular continuous vertex \cite{belhe23} when all adjacent edges are continuous.
The feature $\l_{i, j}, \r_{i, j} \in \R^k$ and the discontinuity weight $w_{i,j} \in \R$ are defined in the same way as the 1D case.
The only difference is how $\l_{i, j}, \r_{i, j}, w_{i,j}$ are defined between adjacent vertices.
The features $\l_{i, j} \neq \l_{j,i}, \r_{i, j} \neq \r_{j,i}$ are defined on the half-edges granting freedom for vertices to have local ``colors'' while the discontinuity weight $w_{i,j} = w_{j,i}$ are shared between two vertices for consistent continuity behavior along an edge.

Like our 1D case, we pass the interpolated feature $\F(\x)$ through a shallow MLP (Algorithm~\ref{alg:inference} \rev{where AngleCCW computes CCW angle between two vectors}).

\begin{algorithm}
  \caption{Field Inference}
  \label{alg:inference}
  \SetKwProg{infer}{Function \emph{infer}}{}{end}
  \infer{($\x$, M)}{
    $T, \lambda_1, \lambda_2 \leftarrow \text{PointInFace}(\x, M)$  \;
    \ForAll{$\v_i$ in $T=(\v_0, \v_1, \v_2)$}{
      \ForAll{$e_i^j$ in $\text{AdjacentHalfEdge}(\v_i)$}{
        $\theta_i^j \leftarrow \text{AngleCCW}((\x - \v_i), e_i^j)$ \;
        $t_i^j \leftarrow \text{fmod}(\frac{\theta - \theta_i^j}{2\pi}+ 1, 1)$\;
      }
      $\hat{\F}_i(\x) \leftarrow \b_i + \sum_{\v_j \in \N(\v_i), i \neq j} \left( w_{i,j} (\l_{i,j} - \r_{i,j}) t_i^j + w_{i,j} \r_{i,j} \right)$ \rev{Eq.~\ref{eq:2d_interp}}\;
    }
    $\F(\x) \leftarrow (1 - \lambda_1 - \lambda_2)\hat{\F}_0(\x) + \lambda_1 \hat{\F}_1(\x) + \lambda_2 \hat{\F}_2(\x)$ \rev{Eq.~\ref{eq:bary}} \;
    \Return $\text{MLP}(\F(\x))$\;
  }
\end{algorithm}

\subsubsection{Comparison to feature space in \cite{belhe23}}

\citeauthor{belhe23}'s per-vertex features are also defined in local polar coordinate
\begin{align}
\label{eq:prev}
\hat{\F}_i = \F_i^{\text{cw}} \frac{\theta_i^{\text{ccw}}}{\theta_i^{\text{cw}} + \theta_i^{\text{ccw}}} + \F_i^{\text{ccw}} \frac{\theta_i^{\text{cw}}}{\theta_i^{\text{cw}} + \theta_i^{\text{ccw}}}.
\end{align}

\begin{figure}[b]
\includegraphics[width=\linewidth]{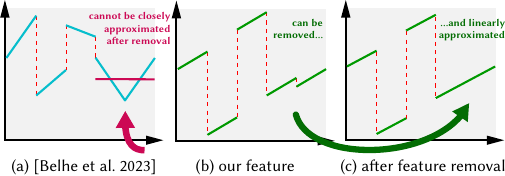}
  \caption{
    (a) Our feature (green) differs from \citeauthor{belhe23}'s feature (cyan).
    (b) This allows us to easily discard almost-continuous edges.
  }
  \label{fig:prevfeature}
\end{figure}

This interpolation scheme (Eq.~2 \cite{belhe23}) finds the closest discontinuous edges clockwise (``cw'') and counter-clockwise (``ccw''), then radially (based on $\theta$s) interpolates the half-edge features $\F_i^{\text{cw}}, \F_i^{\text{ccw}}$ (corresponding to our $\l, \r$ features).
Unlike the entire coverage of $2\pi$ of our interpolation, theirs only covers the domain between two consecutive discontinuities.
Additionally, our local feature function have a constant slope in each piece while theirs can have various slopes (e.g., Fig.~\ref{fig:prevfeature}a).
However, our design is tailored to handle unknown discontinuities.
When a basis vanishes, the edge is continuous, we can easily remove its associated features.
\rev{Compared to ours, \citeauthor{belhe23}'s feature definition allows almost-continuous edges have significantly different slopes in its two adjacent domains.
This difference makes linear approximation (magenta line in Fig.~\ref{fig:prevfeature}a) inaccurate after redundant feature removal, and as a result, \citeauthor{belhe23}'s needs to save more redundant features.}

The disadvantage of having constant piecewise slopes is mitigated by the MLP as experimentally demonstrated in Section~\ref{sec:eval}.
Furthermore, our feature and interpolation scheme can be easily shown to satisfy the continuity criteria proposed by \citeauthor{belhe23} (Section~\ref{sec:method}).
The discontinuous edge set $\Gamma$ is a subset of the mesh edges $E$ and consists of edges where $w_i (\l_{i,j} - \r_{i,j}) \neq \mathbf{0}$.
On each edge, we define the largest feature value jump among all half edges and feature dimensions as $D_{i, j} = \max(w_i(\l_{i,j} - \r_{i,j}))$.
This value, \emph{discontinuity indicator}, reflects the magnitude of the feature space discontinuity.

\subsection{Learning Discontinuous 2D Neural Field}
\label{sec:fit}

\subsubsection{Mesh initialization}
\label{sec:init}

\begin{figure}
  \includegraphics[width=\linewidth]{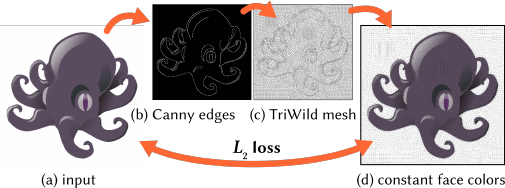}
  \caption{
    We initialize by triangulating Canny edges \cite{canny86} with TriWild \cite{hu19}, then deforming and remeshing interleavedly.
    The deformation is posed as per-face constant color approximation.
  }
  \label{fig:meshing}
\end{figure}

We initialize the triangle mesh $M = (V, T)$ to be roughly aligned with target discontinuities (Fig.~\ref{fig:meshing}).
Note that our method does not require exact edges as it is able to refine edge locations in the next optimization step (Section~\ref{sec:opt}).
But rough alignments are still necessary to obtain discontinuous edges in the early iterations so we can apply differentiable rendering techniques to modify discontinuous edge locations (Fig.~\ref{fig:converge}).

We first roughly detect discontinuities with a Canny edge detector \cite{canny86}. We then connect positive pixels to their corresponding 8-neighbors and apply TriWild \cite{hu19} to generate a triangle mesh $M_0$.
Since the results from the Canny edge detector are inaccurate and limited to the pixel grid, it is only for ensuring the desired triangle density around potential discontinuities.

We use a field $f_0(\x; M)$ with per-face constant colors as a proxy for our initialization.
We approximate the target with $f_0$ by optimizing
\begin{align}
\min_{M} \int_{\Omega} \left\| f_0(\x; M) - I(\x)\right\|^2 d\x  + \lambda_{\text{boundary}} \frac{1}{|\partial M|}\sum_{\v \in \partial M} \left\| \v - \v^0\right\|^2,  
\label{eq:mesh}
\end{align}
where the second term is a MSE loss for softly fixing the boundary \rev{($\partial M$)} to their initial positions $\v^0$s ($\lambda_{\text{boundary}}=10^{-2}$).
We discretize the first term by stratified sampling triangles and compute gradients using SoftRasterizer \cite{liu19}, which can be  replaced by other differentiable renderers.
We interleave the deformation with optional remeshing steps.
See our supplemental document for details.

\subsubsection{Field optimization}
\label{sec:opt}

\begin{figure}
  \includegraphics[width=\linewidth]{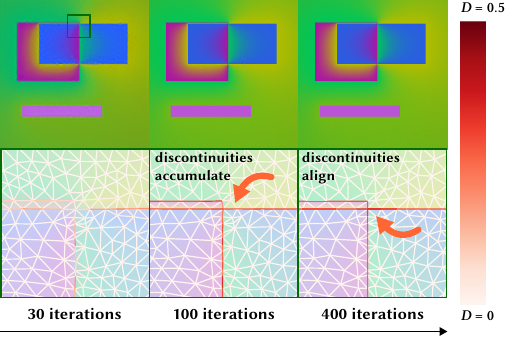}
  \caption{
    We jointly optimize our field and its underlying mesh.
    (a) In the early epochs, our indicated discontinuities begin to accumulate around the target discontinuities.
    (b) As optimization progresses, our method produces close interior color approximation and simultaneously aligns discontinuities.
  }
  \label{fig:converge}
\end{figure}

Given a roughly aligned triangle mesh $M$, we optimize our neural field to approximate the input (Fig.~\ref{fig:converge}).
Our loss function is defined as
\begin{align}
L =  \int_\Omega \left\| f(\x; V, \Theta) - I(\x) \right\|^2 d\x + \lambda_{\text{discont}} \int_{E} \left\| w (\l - \r) \right\|_1 d\x,
\label{eq:fit}
\end{align}
where the first term is a regular $L_2$ fitting loss with varying vertex positions $V$ and neural field parameters $\Theta$ (MLP parameters, feature functions $\F$, and biases $\b$);
the second term is a sparsity inducing term ($\lambda_{\text{discont}}=5\times 10^{-3}$) penalizing feature space value jumps across edges.
We discretize the second term into $\sum_{e_{i,j}\in H} \left\|e_{i,j}\right\| \cdot \left\| w_{i, j} (\l_{i,j} - \r_{i,j})\right\| _1$, where $e_{i,j}\in H$ is a half-edge with length $\left\| e_{i,j}\right\|$, and $w_{i, j}, \l_{i,j}, \r_{i,j}$ are corresponding feature variables.

Note that our loss is similar to that of the MS functional, which contains a fitting term (corresponding to our first term), a smoothness term and a discontinuity sparsifying term (corresponding to our second term).
We do not include an explicit smoothness term and rely on MLP's smoothness bias.

The first term depends on our field function $f$ with discontinuities defined by vertices $V$ and feature functions $\F$.
To correctly estimate the gradient of this integral, we apply edge-sampling Monte Carlo estimation \cite{li18}.
We observe that discontinuity indicator $D = \max(w (\l - \r))$ gives a reasonable estimation about discontinuous edges.
As continuous edges do not contribute to this gradient, we importance sample the discontinuous edges for efficiency.
Although the initial mesh contains edges close to the exact discontinuity locations, the imperfection can result in low discontinuity indication in areas adjacent to true discontinuities.
Because of this, we extend the important edge set to edges with $D > \beta$ and their adjacent edges.
We sample these important edges with $5\times$ probabilities than other edges.
For detailed gradient formulas, see our supplemental.

\subsubsection{Rounding}
\label{sec:round}

\begin{figure}[t]
  \includegraphics[width=\linewidth]{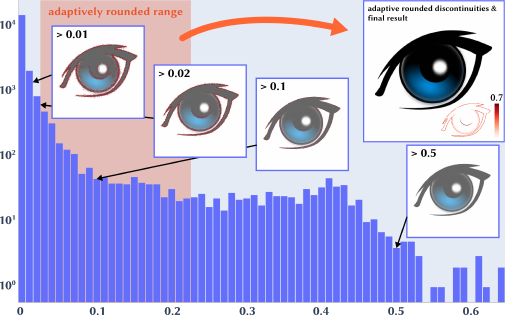}
  \caption{
    \emph{Example log histogram of discontinuity indicator $D$.}
    The majority of edges are identified as continuous, allowing for efficient optimization and storage of our neural field. 
    We apply an adaptive greedily rounding strategy to flexibly determine the final discontinuous edges.
  }
  \label{fig:d_hist}
\end{figure}

Once our neural field sufficiently approximates the input, we discard features on almost-continuous edges.
We refer to this step as ``rounding'', a term borrowed from integer programming \cite{conforti14}.
We only round in one direction---setting $w = 0$ on almost-continuous edges while leaving the remaining $w$s free.

We employ an adaptive rounding strategy (Fig.~\ref{fig:d_hist}).
First, we round by simply thresholding the discontinuity indicator $D$, labeling all edges with $D < \beta$ as continuous.
Then, we greedily discard edges using a priority queue of $D$.
We discard an edge if the accumulated MSE increase is less than $\sigma$ and keep it otherwise.
To discard an edge, we push its contribution $0.5 \cdot w (\l - \r)  + w \r$ to the center vertex bias.
Thanks to the locality of our features, this greedily rounding can be done efficiently.
After the rounding step, we continue refining this new rounded neural field and the mesh with more iterations.

\section{Implementation}

We implemented our method in PyTorch \cite{paszke19}.
To avoid flipped triangles caused by sparse gradients, we apply the large-step precondition \cite{nicolet21} for mesh initialization and field optimization
, with weight of $1$, $0.5$ respectively.
We use ADAM \cite{kingma15} for both steps with $(\beta_1, \beta_2)=(0.9, 0.999)$ and learning rates of $1$ and $2\times10^{-2}$ respectively.

To ensure our $w$s serving as a control over discontinuities, we assign $w = \text{sigmoid}(10 \tilde{w})$ where $\tilde{w}$ is the actual feature parameter.
During the optimization, we use subpixel stratified sampling of $\text{spp} = 2^2$, a key to seamlessly accounting for input anti-aliasing; we set the edge sample number to $4^2 \times W \times H$.
We accumulate gradients from all interior and edge samples per epoch, scheduling our optimization with $70$ epochs of interior fitting and $130$ epochs of interior and edge optimization.
After the rounding step, we continue optimization with both sampling for additional $200$ epochs.

We demonstrate a typical distribution of $D$ (Fig.~\ref{fig:d_hist}).
The threshold $\beta$ defines the hard cutoff of continuous edges and mostly affect the computational cost.
The MSE change threshold $\sigma$ directly affects the result.
We set $\beta=0.02$, $\sigma = 5\times 10^{-6}$ when not explicitly specified.

\section{Evaluation}
\label{sec:eval}

We evaluate our method with various applications: denoising, super-resolution, segmentation and vectorization.
We approximate different types of target functions: constant functions, gradients in vector graphics, harmonic functions in diffusion curves, more complicated functions in synthetic 3D renderings (Fig.~\ref{fig:belhe_comp}), human-drawn artistic images, natural images, and spatial data (depth maps).
We compare to continuous neural fields, represented by a simple reference feature field with per-vertex features and InstantNGP \cite{muller22}, and demonstrate that discontinuity representation is necessary for these tasks.
We evaluate the accuracy of our detected discontinuities by comparing against pixel-based MS functional and a recent triangle-mesh-based MS method \cite{wang22}.
See our supplemental document for setup details.

\begin{figure*}[h!]
  \includegraphics[width=\linewidth]{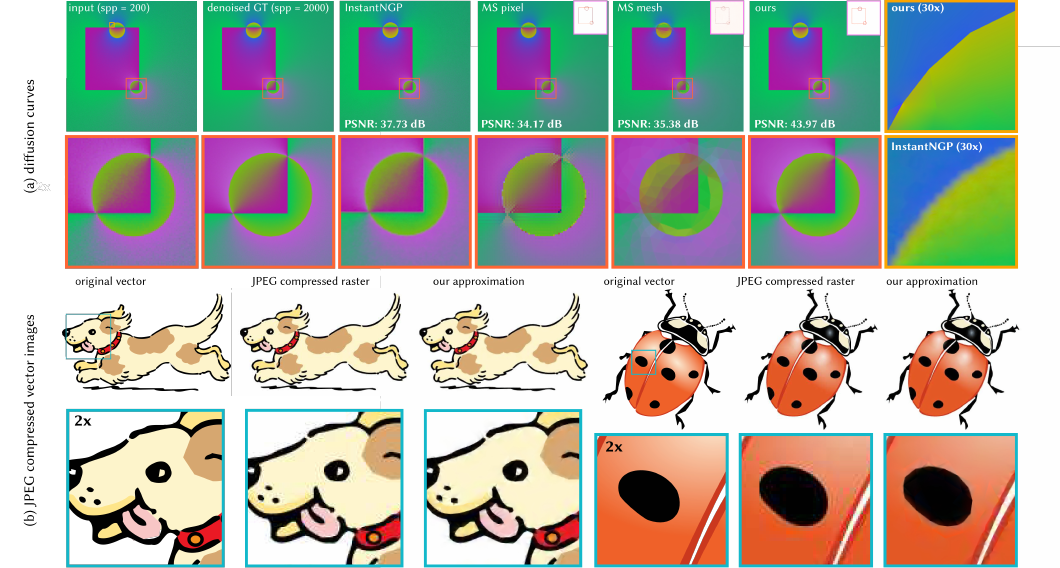}
  \caption{
    \emph{Denoising.}
    (a) We evaluate our model on randomly generated diffusion curves and assess the results for denoising effects and image quality under zoom, compared against continuous neural fields, e.g., InstantNGP \cite{muller22}, and MS functional based methods \cite{wang22}.
    (b) We verify that our neural field can approximate simple constant color fills and gradients under JPEG compression.
  }
  \label{fig:dc_denoise}
\end{figure*}

\paragraph{Denoising}
Noise in images come from many sources.
We test on diffusion curves with noisy Monte Carlo samples rendered using walk-on-spheres (WoS) method \cite{muller56, sawhney20}.
We randomly generate $40$ diffusion curves in the shapes of line segments, rectangles, and circles.
Half of these only contains integer-coordinate rectangles and thus has no anti-aliasing.
These random diffusion curves are rendered with a low number of samples per pixel ($\text{spp}=200$) in $512^2$ resolution and the ground truth (GT) denoised images are generated with $\text{spp}=2000$.
As demonstrated in Fig.~\ref{fig:sec2_comp},\ref{fig:dc_denoise}, our field simultaneously approximates the target harmonic functions and denoises thanks to the smoothness bias of MLP, while the comparison methods either struggles to approximate or over-fits to noise.
We conduct quantitative comparisons (Table~\ref{tab:denoise}), in which our method achieves a significant improvement ($>5$dB) over the second best method, InstantNGP.
Moreover, we verify our method's capability of recovering discontinuities.
We quantify the performance by measuring Chamfer distance between the ground truth diffusion curve \emph{geometries} and the discontinuous \emph{edges} detected by mesh-based MS \cite{wang22} and our method.
Mesh-based MS produces discontinuous edges that are $3.5\times$ Chamfer distance away from the GT compared to our results.
See Fig.~\ref{fig:dc_denoise} for visuals.

Additionally, we qualitatively evaluate our method with the task of denoising JPEG-compressed vector  images (constant color fills and human-created gradients).
Our method not only approximates these two types of functions but also reduces JPEG compression artifacts without prior knowledge about the task.

\begin{table}[t]
  \footnotesize
\caption{
    Quantitative measures.
}
\centering
\label{tab:denoise}
\begin{tabular}{l|rrr|rr}
\hlineB{3}
\multirow{3}{*}{Methods} & \multicolumn{3}{c|}{Denoising} & \multicolumn{2}{c}{\makecell[l]{Denoising +\\ Super-resolution ($2\times$)}} \\
 & PSNR $\uparrow$ & LPIPS $\downarrow$ & \makecell[r]{Median \\ Chamfer $\downarrow$} & PSNR $\uparrow$ & LPIPS $\downarrow$ \\
\hlineB{3}
Per-Vertex & 30.758 & 0.0593 & - & 30.344 & 0.0747  \\
InstantNGP & 39.016 & 0.0431 & - & 32.715 & \textbf{0.0261} \\
\hline
MS Pixel & 35.563 & 0.0319 & - & - & -  \\
MS Mesh & 36.159 & 0.0321 & 0.580 & 36.118 & 0.0466 \\
\hline
Ours & \textbf{44.486} & \textbf{0.0261} & \textbf{0.165} & \textbf{43.913} & 0.0423  \\
\hlineB{3}
\end{tabular}
\end{table}

\paragraph{Super-Resolution}
We qualitatively show zoom-ins ($1.5\times$, $2\times$, $4\times$, $30\times$) of the approximation results across this paper (Fig.~\ref{fig:teaser},\ref{fig:belhe_comp},\ref{fig:sec2_comp},\ref{fig:dc_denoise}).
As presented by the previous work  \citet{belhe23}, although continuous neural fields like InstantNGP closely approximate the input in the original scale, once zoomed in, especially at high zoom levels, they start to exhibit blurs due to lack of discontinuity representation.
We combine the super-resolution quantitative evaluation with the denoising evaluation by measuring the same resulting approximations with $2\times$ zooms.
We measure against the clean diffusion curve images with $\text{spp}=2000$ and $1024^2$ canvas size.
We remark that while the two discontinuity-aware methods, mesh-based MS and ours, output results with similar PSNR as the denoising evaluation, InstantNGP's results experience a drop of PSNR due to the blur artifacts, widening the gap with our method. LPIPS attempts to measure semantic similarity and is less sensitive than PSNR to the absolute function values in the vicinity of discontinuities.

\begin{figure*}
  \centering
  \includegraphics[width=\linewidth]{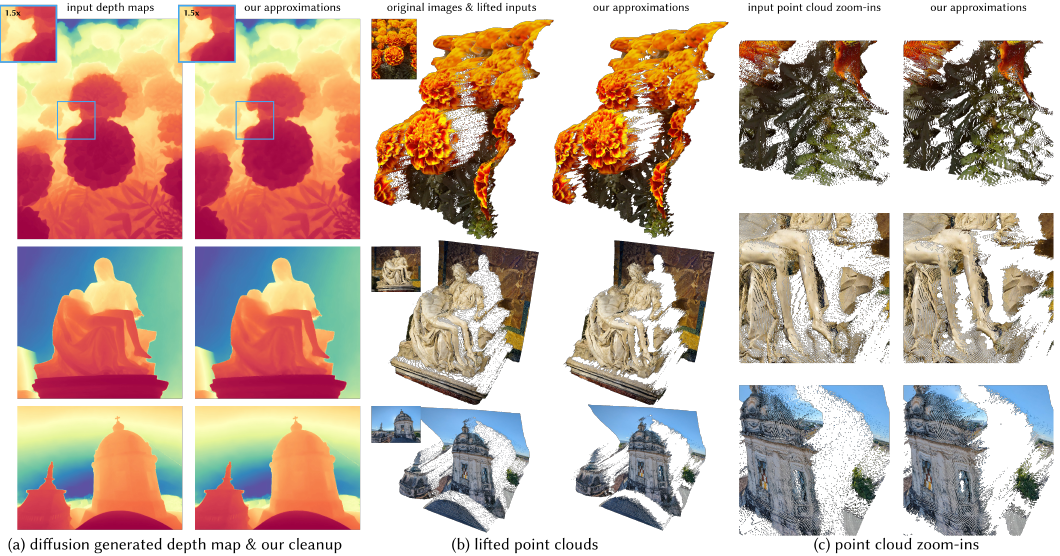}
  \caption{
    \emph{Segmentation of diffusion-generated depth data.}
    (a) Depth map generated by diffusion model \cite{ke23} is corrupted by model-induced noises, particularly around discontinuities.
    (b) Our neural fields closely approximate input data while yielding clean cuts between depth discontinuities.
    See (c) for point cloud zoom-ins.
  }
  \label{fig:segment}

  \vspace*{\floatsep}

  \begin{minipage}{.5\linewidth}
    \includegraphics[width=\linewidth]{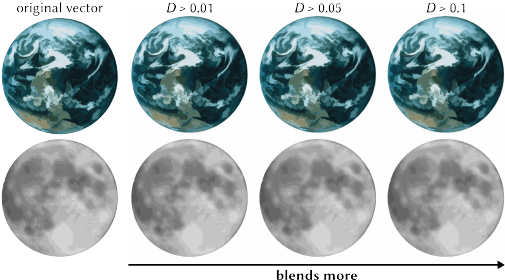}
    \caption{
      \emph{Blending posterized vector images.}
      Vectorization using only solid color fills generates posterized vector images.
      Our method can be applied to blend solid fills and create a resolution-independent image with color gradients.
    }
    \label{fig:blend}
  \end{minipage}%
  \begin{minipage}{.5\linewidth}
  \hspace{\fill}
  \end{minipage}
\end{figure*}

\paragraph{Segmentation}
Beyond RGB(A) images, 2D image functions also include general 2D functions, \emph{spatial data}, such as depth, normal maps, optical flows, and even CLIP feature maps \cite{radford21}.
We evaluate on diffusion-generated depth maps \cite{ke23} (Fig.~\ref{fig:segment}).
Unlike typical scanned depth data, they contain noise originated from neural network, especially around depth discontinuities.
We demonstrate that our method can conduct edge-based segmentation using clean cuts separating depth discontinuities.

\begin{figure*}[h!]
  \includegraphics[width=\linewidth]{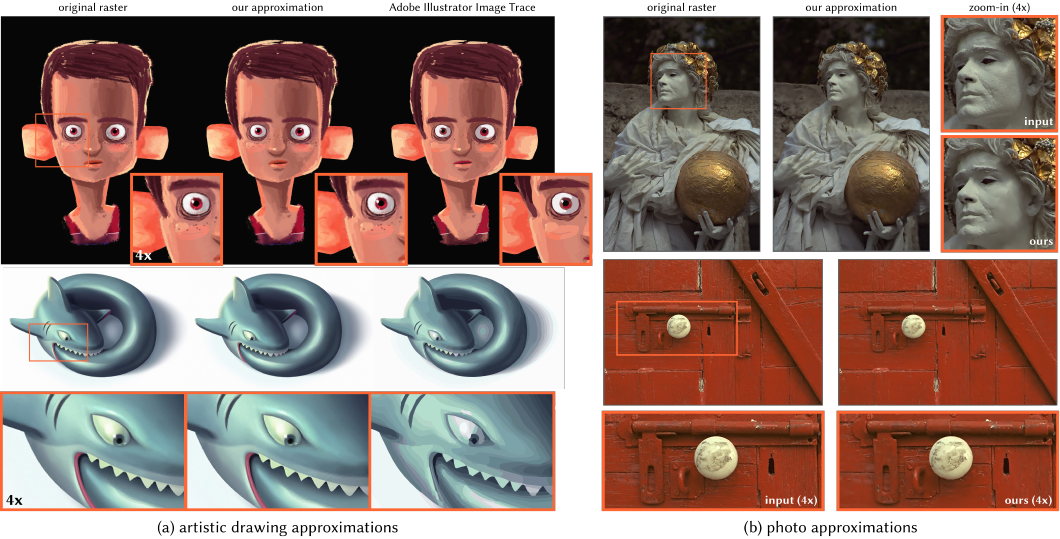}
  \caption{
    \emph{Vectorization of artistic and natural images.}
    (a) Our method qualitatively outperforms common vectorization methods in terms of approximation and representation.
    (b) Our algorithm can vectorize natural images into resolution-independent images with sharp region boundaries when zoomed in.
  }
  \label{fig:vec}
\end{figure*}

\paragraph{Vectorization}
Apart from the synthetic images, we approximate artistic drawings and natural images.
This approximation process can be considered as vectorization since the resulting discontinuous neural fields are resolution-independent, an important property of vector images.
In Fig.~\ref{fig:vec}a, we compare our approximation of artistic inputs to the vectorization of Adobe Illustrator Image Trace, a typical tool using solid fills.
Contrasted with the limited expressiveness of solid fills, our neural field is able to preciously represent the targets and support clear zoom-in views.
As solid-fill vectorization tools, e.g., Adobe Illustrator, Potrace, dominate for this task, we also demonstrate a blending application.
By adjusting the rounding parameters, our neural field can blend solid-fill regions, creating smooth color gradients (Fig.\ref{fig:blend}).
We stress test our method on natural images.
We show in Fig.~\ref{fig:vec}b two successful approximations with mild denoising effects, similar to the edge-preserving bilateral filter.

\section{Limitations and future work}

Our method only supports regular triangle meshes instead of the curved triangle meshes as \citeauthor{belhe23}'s, which remains future work.
We expect this extension to further improve the ability of our neural field to approximate natural images (Fig.~\ref{fig:vec}).
Additionally, our fitting procedure replies on the initial mesh to be reasonably aligned with the target discontinuities.
This may not hold when the target visual feature is small and close to the size of noise or when the colors across the discontinuities are close.
Applying subdivision or remeshing to our neural fields as well as exploring more flexible underlying structures and adapting our proposed feature functions to these structures could be interesting future directions.

Our field is single-level compared to neural fields with multi-level feature grids, such as InstantNGP \cite{muller22}.
This design provides denoising power and compressed field size---sizes of our fields commonly match those of InstantNGPs with $2^{14}$ hash table size (compared to their default $2^{24}$).
Despite these benefits, efficient representation of high-frequency details,
which can stay homogeneous within a continuous region,
is a relevant open question.

\section{Conclusion}
We introduce a mesh-based discontinuous 2D neural field that learns unknown discontinuities, advancing the discontinuity-aware neural fields proposed by Belhe et al. [2023]. 
By treating all mesh edges as potential discontinuities, our method solves for the magnitude of discontinuities through continuous optimization. 
Our approach outperforms continuous neural fields, such as InstantNGP \cite{muller22}, in denoising and super-resolution tasks, maintaining sharp boundaries even at high zoom levels.
Our improvement of \citeauthor{belhe23}'s initial framework provides immediate benefits for applications,including the vectorization of artistic drawings and photos, and the cleanup of diffusion-generated depth data.

\bibliographystyle{ACM-Reference-Format}
\bibliography{00_discontinuity}

\clearpage

\clearpage

\appendix

\section{Implementation Details}

\paragraph{Mesh initialization}
During our mesh initialization, we interleave the deformation with optional remeshing steps.
In a remeshing step, we apply in order
\begin{enumerate}
\item Edge collapses: We remove a face with area smaller than $2\times10^{-5}$ of the canvas area or with an angle greater than $120^{\circ}$ by collapsing its shortest edge.
\item Edge splits: We split a face with $L_2$ fitting loss greater than $L_{\text{split}}=2$ by splitting its longest edge at the mid point.
\item Edge flips: We first flip edges so the triangulation is Delaunay then apply flips to minimize a hybrid loss of $L_2$ fitting loss and Delaunay loss.
\end{enumerate}
The hybrid loss is defined as
\begin{align}
  \sum_{e \in F_i, F_j} \left\| f_0(x; F_i, F_j \in M) - I(x) \right\|^2 + \lambda_{\text{Delaunay}} \text{trace}(\mathbf{L}),
\end{align}
where $F_i, F_j$ are the two faces defined by an edge; $\mathbf{L}$ is the positive semi-definite cotan Laplacian operator defined by the four vertices in $F_i, F_j$; $\lambda_{\text{Delaunay}}$ is set to $0.5$.
It is shown that the trace of cotan Laplacian is decreased by Delaunay flips and reaches a global minimum when the 2D triangulation is Delaunay \cite{alexa19}.

\paragraph{Field optimization}
To jointly optimize our neural field in triangle interiors and the triangle mesh, we compute gradient of the rendering integral with respect to moving triangle area.
This is achieved via common differentiable rendering technique of edge-sampling Monte Carlo estimation \cite{li18}
\begin{align}
&\nabla \int_\Omega \left\|f(\x;V, \Theta) - I(x)\right\|^2 d\x \\
&= \left(\nabla^T \int_\Omega  f(\x;V, \Theta)  d\x\right) \cdot \int_\Omega (f(\x;V, \Theta) - I(\x)) d\x.
\end{align}

This gradient can be estimated as
\begin{align}
& \nabla\int_\Omega  f \, dx \\
= &\int_{\Omega \setminus E} \frac{\partial}{\partial \Theta} f \, d\x + \sum_{e_i \in E}\int_{\alpha_i(\x)=0} \frac{\nabla_V \alpha_i(\x)}{\left\|\nabla_V \alpha_i(\x)\right\|} f(\x) \, dp(\x),
\end{align}
where the first term is the regular gradient of the field with respect to the field parameter in the face interiors; the second term is the gradient with respect to the vertices along discontinuous edges.
The edges in the second term is defined by implicit functions $\alpha(x)$.

\section{Parameter Configurations}

\paragraph{Mesh Initialization}
We detect Canny edges with low and high thresholds of $100$ and $200$.
To reduce the salt-and-pepper noises in the 3D renderings, we smooth the image with Gaussian kernel of size $3$ before Canny edge detection.
We call TriWild with target edge length ratio of $10^{-2}$.
We deform the mesh with SoftRasterizer (sharpness kernel size of $10^{-1}$) for $200$ epochs and remesh every $50$ epochs.
The per-face colors are set to be the face mean color in every iteration rather than kept as variables.
Our remeshing implementation is based on the continuous remeshing PyTorch implementation \cite{palfinger22}.
We employ per-triangle stratified sampling to avoid flipping very small triangles during mesh initialization.
For the artistic drawing inputs where noise is minimal and faithful approximation is the goal, the TriWild target edge length ratio is adjusted to $3\times 10^{-3}$ and the mesh deformation is run for 100 epochs without remeshing.

\paragraph{Field Optimization}
We initialize all our neural field weights using standard Xavier normal distribution and all biases as zeros.

\section{Comparison Setup}

\paragraph{Rendering}
We render the resolution-independent results of mesh-based MS, InstantNGP, and our method using subpixel grid samples ($\text{spp} = 4^2$) for better visual effects given the target resolution.

\paragraph{JPEG compressed vector images}
We generate input raster image by normalizing a vector image such that its longer axis occupies $90\%$ of the $512\times 512$ white canvas, then rasterizing the image with \citeauthor{inkscape} and compressing it into a JPEG image with quality of $50$ via \citeauthor{imagemagick}.

\paragraph{InstantNGP}
We use the tiny-cuda-nn \cite{tiny-cuda-nn} based implementation for subpixel inference.
We match the size of InstantNGP and our neural field by adjusting InstantNGP's hash table size.
We ensure InstantNGP's convergence by running for $10k$ iterations.

\paragraph{Mumford-Shah functional}
We reimplement a mesh-based MS denoising algorithm \cite{wang22} and run it on the resulting aligned mesh from our method.
The pixel-based MS denoising algorithm is implemented by replacing \citeauthor{wang22}'s discretization with the one on pixel grids.
We conduct experiments with parameters: $\alpha = 1, \beta = 0.01, \gamma = 100, epsilon = 0.01$ and $10$ iterations of alternating optimization.
For comparison fairness, we grid search the discontinuity threshold of mesh-based MS in the range of $v = [0.01, 0.09]$ (step size of $0.01$) and $[0.1, 0.5]$ (step size of $0.1$) given the Chamfer distance.

\paragraph{Field of Junctions}
We use the official implementation of FofJ \cite{verbin21}.
We manually tune the parameters for the most accurate approximation.
Our experiment uses $\eta = 0.01$, $\delta = 0.1$, patch size $R = 17$, stride $s = 5$ for a $512\times 512$ image, consistency weights $\lambda_B = 0.5$, $\lambda_C = 0.1$, $N_{init} = 30$ iterations, $N_{iter} = 1000$ iterations, the number of values to query $N = 10$ in Algorithm 2, learning rate of 0.03 for the vertex positions and 0.003 for the junction angles in the refinement step to generate the global boundaries and smoothed image.

\section{Ablations}

\begin{figure}
\includegraphics[width=\linewidth]{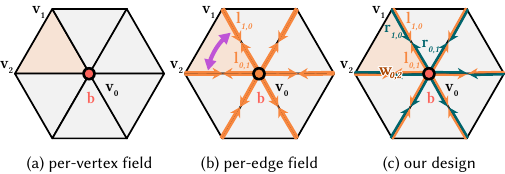}
  \caption{
    Different designs of mesh-based feature fields.
  }
  \label{fig:ablatione}
\end{figure}

\begin{table}[t]
  \footnotesize
\caption{
    Ablation of feature design.
}
\centering
\label{tab:ablation}
\begin{tabular}{l|rr|rr}
\hlineB{3}
\multirow{3}{*}{Methods} & \multicolumn{2}{c|}{Denoising} & \multicolumn{2}{c}{\makecell[l]{Denoising +\\ Super-resolution ($2\times$)}} \\
 & PSNR $\uparrow$ & LPIPS $\downarrow$  & PSNR $\uparrow$ & LPIPS $\downarrow$ \\
\hlineB{3}
Per-Vertex & 30.758 & 0.0593 & 30.344 & 0.0747  \\
Per-Edge & 36.682 & \textbf{0.0178} & 35.518 & \textbf{0.0361}  \\
\hline
Ours & \textbf{44.486} & 0.0261  & \textbf{43.913} & 0.0423  \\
\hlineB{3}
\end{tabular}
\end{table}

\begin{figure}
\includegraphics[width=\linewidth]{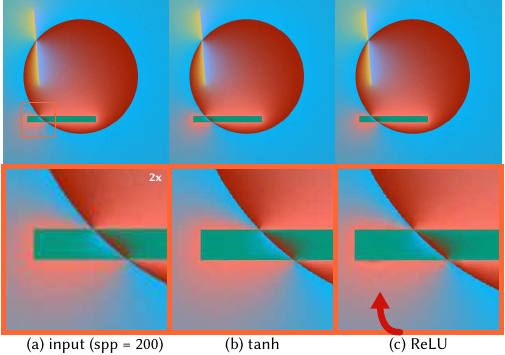}
  \caption{
    ReLU activations generate redundant discontinuities.
  }
  \label{fig:relu}
\end{figure}

\paragraph{Feature design}
We compare to two simpler versions of our neural field: per-vertex field and per-edge field (Fig.~\ref{fig:ablatione}).
The per-vertex field is a hybrid feature field with features stored at vertex.
We show and report the results of this per-vertex field as a reference in the main text.
We ablate our design of discontinuous edges by introducing a per-edge field.
This field interpolates the features stored on the two consecutive half-edges of the query face (purple arrows in Fig.~\ref{fig:ablatione}).
\rev{This interpolation is similar to our method but the feature of this configuration is continuous across any edges.}
Both versions are significantly worse than our model as reported in Table~\ref{tab:ablation} and shown in our supplementary materials.
Similar to the InstantNGP results, we observe that LPIPS is less sensitive to the absolute function values in the vicinity of discontinuities.

\paragraph{Activation}
We choose tanh for activations since we observe that ReLU compete with discontinuous edges.
As utilized by ReLU fields \cite{karnewar22}, ReLU activations tend to create steep slopes, serving an overlapping role as our discontinuous features.
In Fig.~\ref{fig:relu}, our neural field with ReLU activations is less smooth compared to the version with tanh activations.
As we reply on the smoothness of MLP, we decide on tanh for activations.

\end{document}